%% file: main.tex
\pdfoutput=1

\documentclass[11pt]{article}

\usepackage[]{acl}

\usepackage{times}
\usepackage{latexsym}

\input{header}
\input{math_commands}

\usepackage[T1]{fontenc}

\usepackage[utf8]{inputenc}

\usepackage{microtype}

\title{Sentiment Analysis through LLM Negotiations}

\author{Xiaofei Sun$^{\blacklozenge}$, Xiaoya Li$^{\clubsuit}$, Shengyu Zhang$^{\blacklozenge}$, Shuhe Wang$^{\blacktriangledown}$
\\ \bf Fei Wu$^{\blacklozenge}$, 
Jiwei Li$^{\blacklozenge}$, Tianwei Zhang$^{\heart}$, Guoyin Wang$^{\clubsuit}$}

\begin{document}

\maketitle
\begin{abstract}
A standard paradigm for sentiment analysis is to 
rely on a singular LLM and makes the decision in a single round under the framework of in-context learning.
This framework suffers the key disadvantage that 
the single-turn output generated by a single LLM might not deliver the perfect decision, just as humans sometimes need multiple attempts to get things right. 
This is especially true for the task of sentiment analysis where deep
reasoning 
is required to address 
 the complex linguistic phenomenon (e.g., clause composition, irony, etc) in the input.

To address this issue, this paper introduces a multi-LLM negotiation framework 
 for sentiment analysis.
 The framework consists of  a reasoning-infused generator to provide decision along with rationale, 
 a explanation-deriving discriminator to evaluate the credibility of the generator.
 The generator and the discriminator iterate until a consensus is reached. 
The proposed framework naturally addressed the aforementioned challenge, as we are able to take the complementary abilities of two LLMs, 
have them use rationale to persuade each other for correction.  

Experiments on a 
wide range of
 sentiment analysis 
benchmarks (SST-2, Movie Review, Twitter, yelp, amazon, IMDB)
demonstrate 
the effectiveness of proposed approach: 
it 
consistently
yields better performances 
than the ICL baseline across all benchmarks, and even superior performances to  supervised baselines on the Twitter and movie review datasets.
\noindent\let\thefootnote\relax\footnotetext{$^{\blacklozenge}$Zhejiang University,$^{\clubsuit}$Shannon.AI,$^{\blacktriangledown}$Peking University, $^{\clubsuit}$Bytedance,$^{\heart}$Nanyang Technological University}
\noindent\let\thefootnote\relax\footnotetext{\{xiaofei\_sun, sy\_zhang, wufei, jiwei\_li\}@zju.edu.cn, xiaoya\_li@shannonai.com, wangshuhe@stu.pku.edu.cn, tianwei.zhang@ntu.edu.sg, guoyin.wang@bytedance.com}
 \end{abstract}

\section{Introduction}
Sentiment analysis~\citep{Pang2008OpinionMA,  go2009twitter, maas2011learning, Zhang2012SentimentAA, Baccianella2010SentiWordNet3A, Medhat2014SentimentAA, Bakshi2016OpinionMA, Zhang2018DeepLF} 
aims to 
extract opinion polarity expressed by a chunk of text.  
Recent advances in large language models (LLMs)~\citep{brown2020language, ouyang2022training, touvron2023llama, touvron2023llama2, anil2023palm, zeng2022glm,  OpenAI2023GPT4TR, bai2023qwen} open a new door
 for the resolving the task~\citep{lu2021fantastically, kojima2022large, wang2022self, wei2022chain, wan2023gpt, wang2023gpt, sun2023text, sun2023pushing, lightman2023let, li2023human, schick2023toolformer}:
under the paradigm of in-context learning (ICL), 
LLMs are able to achieve performances comparable to supervised learning strategies~\citep{lin2021bertgcn, sun2021chinesebert, Phan2020ModellingCA, Dai2021DoesSM} with only a small number of training examples. 



Existing approaches that harness LLMs for sentiment analysis usually rely on  a \textbf{singular LLM}, and make a decision in a \textbf{single round} under ICL.
This strategy suffers from the following disadvantage:
the single-turn output generated by a single LLM  might not deliver the perfect response:
Just as humans sometimes need multiple attempts to get things right,  
it might take multiple rounds before an LLM makes the right decision.
This is especially true for the task of sentiment analysis, where LLMs usually need to  articulate the 
reasoning process to address the  
complex linguistic phenomenon (e.g.,
clause composition,  irony, etc) in the input sentence.



\begin{figure}[!t]
 \centering
 \begin{minipage}[l]{0.95\linewidth}
\hspace{-0.3cm}
\includegraphics[scale=0.22]{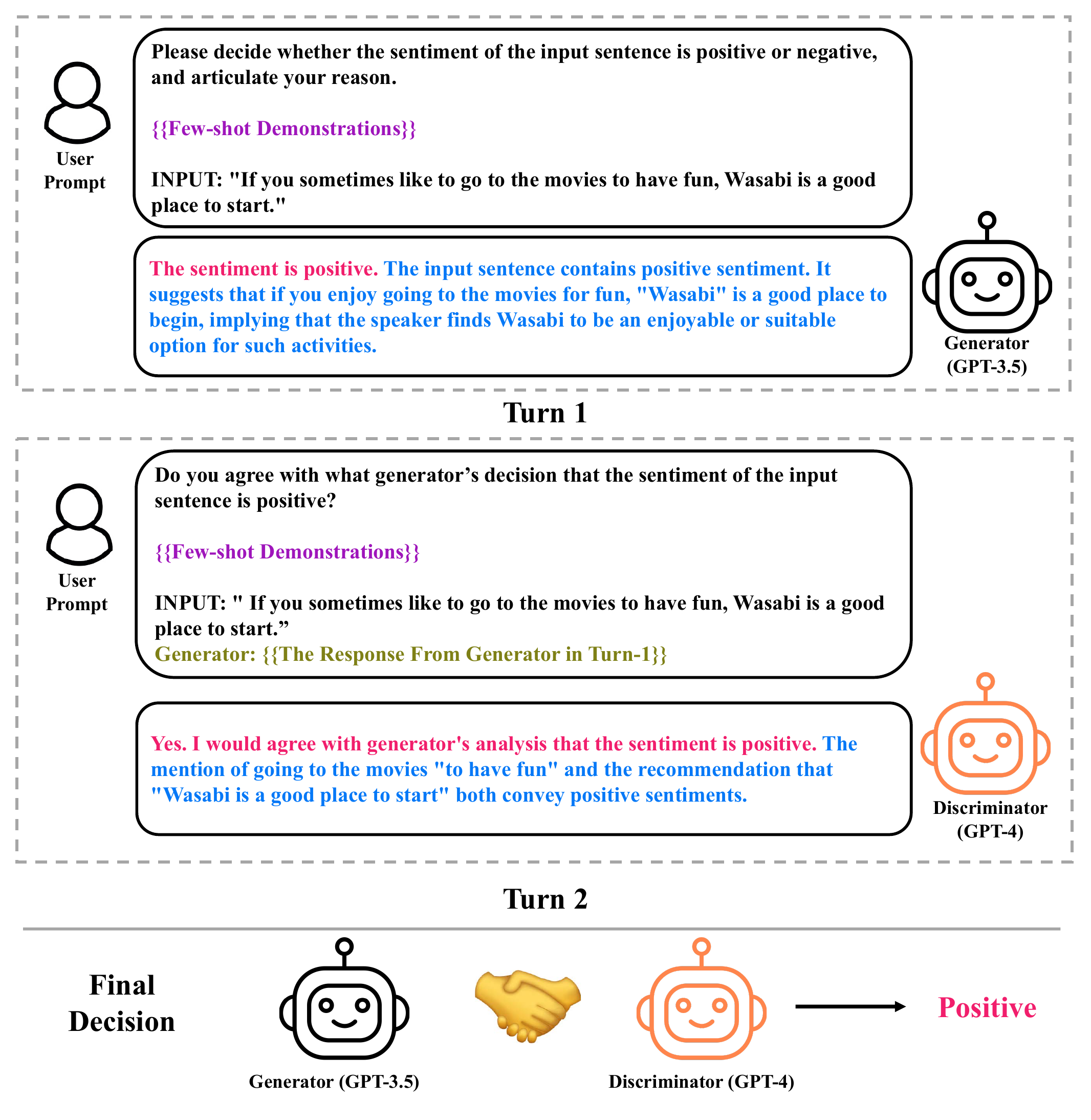}
\caption{An illustration of a generator (G) and a {\color{discriminator} discriminator (D)} achieving consensus via a negotiation. Each round consists of a user prompt and a response from either G or {\color{discriminator} D}. Specifically, a user prompt includes four elements: a task description, {\color{demo}  few-shot demonstrations } (abbreviate it for short), an input, and a {\color{response} response from the last turn} (if applicable). Responses from G or {\color{discriminator} D} start with statements that the input contains {\color{positive} positive} sentiment, followed by {\color{rationale} rationale}. } 
\label{fig:sample_figure}
 \end{minipage}
 \end{figure}

To address the 
this issue, 
in this paper,
we propose a multi-LLM negotiation strategy  
for  sentiment analysis. 
The core of the proposed strategy 
is a generator-discriminator  framework,
where 
one LLM acts as the generator (G) to produce sentiment decisions, while the other acts as a discriminator (D), tasked with evaluating the credibility  of the generated output from the first LLM. 
The proposed method innovates on three aspects:  (1) Reasoning-infused generator (G):
an LLM that
adheres to a structured reasoning chain, enhancing the ICL of the generator while offering the discriminator
the
 evidence and insights
to evaluate its validity; 
(2) Explanation-deriving discriminator (D);
other LLM 
designed to offer post-evaluation rationales for its judgments;
(3) Negotiation: 
 two LLMs act as the roles of the generator and the discriminator,   and 
 perform the negotiation 
  until a consensus is reached.  
  
This strategy harnesses the collective abilities of the two LLMs and provide with the channel for the model to correct imperfect responses,
and thus naturally resolves the issue that a single LLM cannot yield the correct decision on its first try. 



The contributions of this work can be summarized as follows:
1) we provide a novel perspective on how sentiment analysis can benefit from multi-LLM negotiation. 2) we introduce a Generator-Discriminator Role-switching Decision-Making framework that enables multi-LLM collaboration through iteratively generating and validating sentiment categorizations. 3) our empirical findings offer evidence for the efficacy of 
the proposed
 approach:
 experiments on a 
wide range of
 sentiment analysis 
benchmarks (SST-2, Movie Review, Twitter, yelp, amazon, IMDB)
demonstrate 
that 
the proposed method
consistently
yields better performances 
than the ICL baseline across all benchmarks, and even superior performances to  supervised baselines on the Twitter and movie review datasets.

\section{Related Work}

\subsection{Sentiment Analysis}

Sentiment analysis~\citep{Pang2008OpinionMA, go2009twitter, maas2011learning, Zhang2012SentimentAA, Baccianella2010SentiWordNet3A, Medhat2014SentimentAA, Bakshi2016OpinionMA, Zhang2018DeepLF} is a task that aims to determine the overall sentiment polarity (e.g., positive, negative, neutral) of a given text. Earlier work often formalized the task as a two-step problem: (1) extract features using RNNs~\citep{socher2013recursive, qian2016linguistically, peled2017sarcasm, wang2016combination, Guggilla2016CNNAL, vo2015target}, CNNs~\citep{kalchbrenner2014convolutional, wang2016dimensional, Guan2016WeaklySupervisedDL, Yu2016LearningSE, Mishra2017LearningCF},  pretrained language models~\citep{lin2021bertgcn, sun2021chinesebert, Phan2020ModellingCA, Dai2021DoesSM}, etc; and (2) feed extracted features into a classifier for obtaining a pre-defined sentimental label. 

In recent years, in-context learning (ICL) has achieved great success and changed the paradigm of NLP tasks. 
Many works adapt ICL to the sentiment analysis task:
\citet{qin2023chatgpt, sun2023pushing} propose a series of strategies to improve ChatGPT's performance on the sentiment analysis task;
\citet{fei2023reasoning} propose a three-hop reasoning framework, which induces the implicit aspect, opinion, and finally the sentiment polarity for the implicit sentiment analysis task;ƒ
\citet{zhang2023sentiment} find that LLMs can achieve satisfactory performance on the binary sentiment classification task, but they underperform to the supervised baseline on more complex tasks (e.g., fine-grained sentiment analysis) that require deeper understanding or structured sentiment information.

\begin{figure*}[!ht]
 \centering
 \begin{minipage}[c]{0.92\linewidth}
\vspace{-1.9cm}
\hspace{0.13cm}
\includegraphics[scale=0.20]{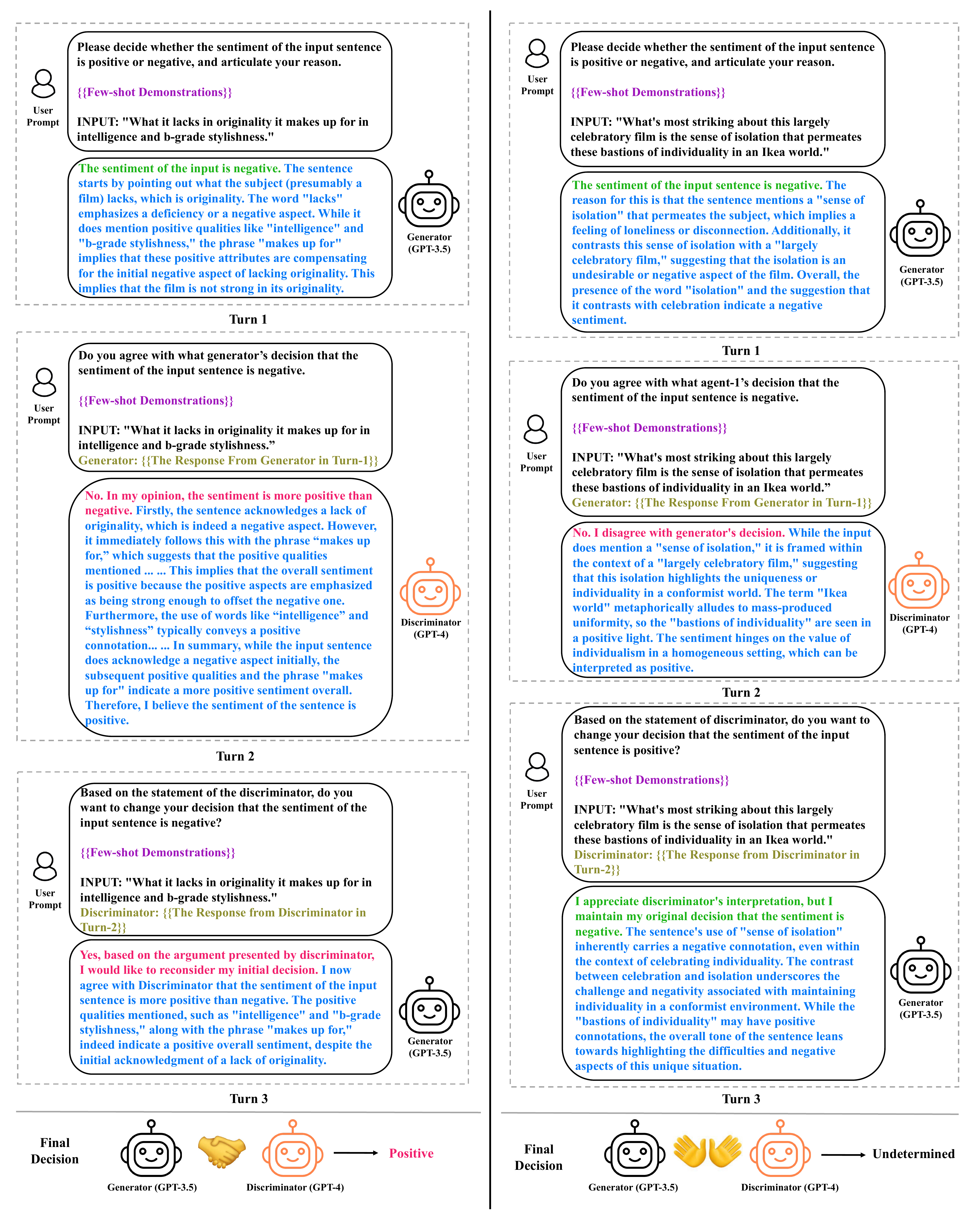}
\caption{Illustration of the negotiation procedure. The left demonstration shows a case where an agreement on the positive sentiment is reached after turns turns, while the right demonstration shows a case where two LLMs fail to reach an agreement in three turns. Specifically, a user prompt includes four elements: a task description, {\color{demo}  few-shot demonstrations } (abbreviate for short), an input, and a {\color{response} response from the last turn} (if applicable). Responses from the generator or {\color{discriminator} discriminator} start with statements that the input contains either {\color{positive} positive} or {\color{negative} negative} sentiment, followed by {\color{rationale} rationales}.}
\label{fig:sample_figure2}
 \end{minipage}
 \end{figure*}

\subsection{Large Language Models and In-context Learning}
Large language models (LLMs)~\citep{wang2022pre, zhang2023instruction} are models trained on massive unlabeled text corpora with self-supervised learning techniques. Based on the model architecture, LLMs can be categorized into three types: (1) encoder-only models, which contain a text encoder and generate the input representations, such as BERT~\citep{devlin2018bert} and its variants~\citep{lan2019albert, liu2019roberta, sun2020ernie, clark2020electra, feng2020codebert, joshi2020spanbert, sun2020ernie, sun2021chinesebert}; (2) decoder-only models, which have a decoder and generate text conditioned on the input text like GPT-series models~\citep{radford2019language, brown2020language, keskar2019ctrl, radford2019language, chowdhery2022palm, ouyang2022training, zhang2022opt, scao2022bloom, zeng2022glm, touvron2023llama, peng2023instruction, OpenAI2023GPT4TR}; and (3) encoder-decoder models, which have a pair of encoder-decoder and generate text conditioned on the input representation, such as T5~\citep{raffel2020exploring} and its variants~\citep{lewis2019bart, xue2020mt5}.

Starting with GPT-3~\citep{brown2020language}, LLMs have shown emerging capabilities~\citep{wei2022emergent} and completed NLP tasks through in-context learning (ICL), where LLMs generate label-intensive text conditioned on a few annotated examples without gradient updates. 
Many studies in the literature propose strategies for improving ICL performances on NLP tasks. 
\citet{li2021prefix, chevalier2023adapting, mu2023learning} optimize prompts in the continuous space. 
\citet{liu2021makes, wan2023gpt, zhang2023promptner} search through the train set to retrieve $k$ nearest neighbors of a test input as demonstrations. 
\citet{zhang2022automatic, sun2023text, yao2023tree} decompose a task into a few sub-tasks and solve them step-by-step towards the final answer conditioned on LLM-generated reasoning chains.
\citet{sun2023pushing, wang2023gpt} propose to verify LLMs' results by conducting a new round of prompting;
\citet{liu2021generated, Feng2023FactKBGF} use LLMs to generate natural language knowledge statements and integrate external knowledge statements into prompts. 

\subsection{The LLM collaboration}
The LLM collaboration involves multiple LLMs working together to solve a given task. Specifically, the task is decomposed to several intermediate tasks, and each LLM is assigned to complete one intermediate task independently. The given task is solved after integrating or summarizing these intermediate results. The LLM collaboration approach can exploit the capabilities of LLMs, improve performances on complex tasks and enable to build complicated systems. \citet{Shinn2023ReflexionAA, sun2023pushing, gero2023selfverification, Wang2023LearnFM, Chen2023TeachingLL} construct auxiliary tasks (e.g., reflection, verification tasks) and revise the response to the original task referring to the result of the auxiliary task. \citet{talebirad2023multi, hong2023metagpt, qian2023communicative} assign characterize profiles (e.g., project manager, software engineer) to LLMs and gain performance boosts on character-specific tasks through behavior animations. \citet{li2022composing, zeng2022socratic, chen2023reconcile, du2023improving, Liang2023EncouragingDT} use a debate strategy in which multiple different LLMs propose their own responses to the given task and debate over multiple turns until getting a common final answer. Besides, \citet{shen2023hugginggpt, gao2023assistgpt, ge2023openagi, zhang2023automl, hao2023chatllm} employ one LLM as the task controller, which devises a plan for the given task, selects expert models for implementation and summarizes the responses of intermediate planned tasks. Other LLMs serve as task executors, completing intermediate tasks in their areas of expertise.

\section{LLM Negotiation for Sentiment Analysis}
\subsection{Overview}
In this section, we detail 
the
multi-LLM negotiation framework for sentiment analysis:
Two LLMs perform as the answer generator and discriminator. We refer to the interaction between the generator and the discriminator as a negotiation. 
The negotiation will repeat until a consensus is reached or the maximum number of negotiation turns is exceeded.
Illustrations are shown in Figures \ref{fig:sample_figure} and \ref{fig:sample_figure2}.  

\subsection{Reasoning-infused generator}
The  generator is backboned by a large language model. We ask the answer generator based on the ICL paradigm through prompts, aiming to generate a step-by-step reasoning chain and a decision towards the sentiment polarity of the test input. 

Prompts are composed of three elements: a task description, demonstrations, and a test input. The task description is a description of the task in natural language (e.g., "Please determine the overall sentiment of test input."); the test input is the textual input in the test set (e.g., "The sky is blue."); demonstrations are from the train set of the task. Each consists of three elements: input, reasoning chains, and sentimental decision.

For each test input, we first retrieve $K$ nearest neighbors (input, sentiment decision) from the train set as demonstrations. Then, we transform demonstrations to (input, reasoning process, sentiment decision) triplets by prompting the generator to produce a reasoning chain. After concatenating the task description, demonstrations, and the test input, we forward the prompt to the generator, which will respond with a step-by-step reasoning chain and a sentimental decision. 

\subsection{Explanation-deriving discriminator}
The  discriminator is backboned by another LLM. After finishing the answer generating process, the answer discriminator is used to judge whether the decision made by the generator is correct and provide a reasonable explanation.

To accomplish this goal, we first construct prompts for the answer discriminator. The prompt is composed of four elements: a task description, demonstrations, a test input, and the response from the answer generator. The task description is a piece of text that describes the task in natural language (e.g., "Please determine whether the decision is correct."). Each demonstration is composed of six elements: (input text, a reasoning chain, sentiment decision, discriminator attitude, discriminator explanations, discriminator decision) and constructed by prompting the answer discriminator to provide explanations of why the sentiment decision is correct for the input text.

Then we ask the discriminator with the construct prompt. The answer discriminator will respond with a text string, containing an attitude (i.e., yes, no) that denotes whether the discriminator agrees with the generator, explanations that explain why the discriminator agrees/disagrees with the generator, and a discriminator decision that determines the sentiment of the test input.

\paragraph{Why Two LLMs but Not One?}
There are two reasons for using two different LLMs separately for the generator and the discriminator rather than using a single LLM to act as two roles: (1)
If an LLM makes a mistake as a generator due to incorrect reasoning, it is more likely that it will also make the same mistake as the discriminator as since 
generator and the discriminator from the same model are very likely to make similar rationales; (2) by using two separate models, we are able to take the advantage 
of the complementary abilities of the two models. 


\subsection{ Role-flipped Negotiation}
After two LLMs end with a negotiation, 
we ask them flip roles and initiate a new negotiation,
where the second LLM acts as the generator, and the first LLM acts as the discriminator. 
 We refer the interaction of two LLMs with flipped roles as role-flipped negotiation. Likewise, the role-flipped negotiation is ended until a consensus is reached or the maximum number of negotiation turns is exceeded.

 When both negotiations result in an agreement and their decisions are the same, we can choose either decision as the final one since they are the same. If one of the negotiations fails to reach a consensus while the other reaches a decision, we choose the decision from the negotiation that reached a consensus as the final decision. However, if both negotiations reach a consensus but their decisions do not align, we will require the assistance of an additional Language Model (LLM), as will be explained in more detail below."

\paragraph{Introducing a third LLM } 
If the decision from the two negotiations do not align,
 we introduce a third LLM and conduct the negotiation and role-flipped negotiation with each of the two aforementioned LLMs. Subsequently, we will get 6 negotiation results and vote on these results: the decision that appears most frequently is taken as the sentiment polarity of the input test.

\section{Experiments}
To evaluate the effectiveness of the proposed
method, we use GPT-3.5, GPT-4~\citep{OpenAI2023GPT4TR} and InstructGPT3.5~\citep{ouyang2022training}\footnote{\texttt{text-davinci-003}} as backbones for the multi-model negotiation method. In this process, we use the fine-tuned RoBERTa-Large~\citep{liu2019roberta} as the similarity function for retrieving $k$ nearest neighbors as demonstrations. 

In the empirical study, we investigate the following three distinct ICL approaches, offering insights of integrating such methods for sentiment analysis.
\begin{tightitemize}
    \item {\bf Vanilla ICL}: the sentiment analysis task is finished by asking a LLM with a prompt to generate sentiment-intensive text without gradient updates. In practice, we conduct two sets of experiments under this setting with GPT3.5 and GPT-4, respectively. 
    \item {\bf Self-Negotiation}: the task is finished by using one LLM to discriminate and correct the answer generated by itself. We conduct two experiments with GPT3.5 and GPT-4 and get two results.
    \item {\bf Negotiation with two LLMs}: the task is completed by employing two different LLMs to take turns performing as the answer generator and discriminator. Specifically, we conduct one set of experiment with GPT3.5 and GPT-4. 
\end{tightitemize}


\subsection{Datasets}
We conduct experiments on six sentiment analysis datasets, including SST-2~\citep{socher2013recursive}, Movie Review~\citep{zhang2015character}, Twitter~\citep{rosenthal2019semeval}, Yelp-Binary~\citep{zhang2015character}, Amazon-Binary~\citep{zhang2015character}, and IMDB~\citep{Maas2011LearningWV}. More details of the datasets are shown as follows: 
\begin{tightitemize}
\item {\bf SST-2}~\citep{socher2013recursive}: SST-2 is a binary (i.e., positive, negative) sentiment classification dataset and contains movie review snippets from the Rotton Tomato. We follow \citet{socher2013recursive} and use the train, valid, test splits with the number of examples of 67,349, 872, 1,821, respectively.
\item {\bf Movie Review (MR)}~\citep{zhang2015character}: Movie Reviews is a dataset for use in sentiment-analysis experiments. Available are collections of movie-review documents labeled with respect to their overall sentiment polarity (i.e., positive or negative).
\item {\bf Twitter}~\citep{rosenthal2019semeval}: Twitter is a three-class (i.e., positive, negative, neutral) sentiment analysis dataset, aiming to detecting whether a piece of text expresses a sentiment polarity in respect to a specific topic, such as a person, a product, or an event. The dataset is origin a shared task at SemEval 2017, containing 50,333 examples in the train set and 12,284 examples in the test set. 
\item {\bf Yelp-Binary}~\citep{zhang2015character}: Yelp is a binary (i.e., positive, negative) sentiment analysis dataset, containing product reviews from Yelp. The dataset has 560,000 trainig samples and 38,000 testing samples. 
\item {\bf Amazon-Binary}~\citep{zhang2015character}: Amazon is a binary sentiment classification task, containing product reviews from Amazon with 3,600,000 examples in the train set and 400,000 examples in the test set. 
\item {\bf IMDB}~\citep{Maas2011LearningWV}: The IMDB dataset contains movie reviews along with their associated binary sentiment polarity labels. 
The dataset contains 50,000 reviews split evenly into 25k train
and 25k test sets. The overall distribution of labels is balanced (25k
positive and 25k negative). 
\end{tightitemize}

We use accuracy as the evaluation metric. 

\input{tables/main_result}

\subsection{Baselines}
We use supervised neural network models and ICL approaches with LLMs as baselines for comparisons.
For supervised methods, we choose the following four models: 
\begin{tightitemize}
    \item {\bf DRNN}~\citep{wang2018disconnected}: incorporates position-invariance into RNN and CNN models by limiting the distance of information flow in neural networks. 
    \item {\bf RoBERTa}~\citep{liu2019roberta}: is a reimplementation of BERT~\citep{devlin2018bert} aiming to improve performances on NLP downstream tasks. In this paper, we report results achieved by fine-tuned RoBERTa-Large.
    \item {\bf XLNet}~\citep{yang2019xlnet}: is a pre-trained autoregressive LM that integrates Transformer-XL~\citep{dai2019attentive} and enables to learn bidirectional contexts by maximizing the expected likelihood over all permutations of the factorization order.
    \item {\bf UDA}~\citep{xie2020unsupervised}: is short for Unsupervised Data Augmentation, which is a data augmentation strategy that employs a consistency loss function for unsupervised and supervised training stages. Performances in Table~\ref{tab:main-results} are obtained by BERT-Large with UDA.
    \item {\bf BERTweet}~\citep{nguyen2020bertweet}: is a pre-trained language model for English Tweets. The BERTweet has the same number of parameters as RoBERTa-Base.
    \item {\bf EFL}~\citep{wang2021entailment}: is backboned by RoBERTa-Large and fine-tuned on natural language entailment examples. 
\end{tightitemize}

For ICL approaches, we report experimental results with LLMs from the following studies:
\begin{tightitemize}
    \item {\citet{zhang2023sentiment}}: presents a comprehensive study for applying LLMs (i.e., FLan-UL2, T5 and ChatGPT) on sentiment analysis tasks. Experimental results in the Table~\ref{tab:main-results} are obtained in few-shot($k=5$) settings.
    \item {\bf InstructGPT-3.5}~\citep{ouyang2022training}: is a large language model trained to follow human instructions. Experimental results in the Table~\ref{tab:main-results} are achieved by the \texttt{text-davinci-003} model.
    \item {\bf IDS}~\citep{Qin2023InContextLW}: propose an Iterative Demonstration Selection (IDS) strategy to select demonstrations from diversity, similarity, and task-specific perspectives. Results shown in Table~\ref{tab:main-results} are obtained by using GPT-3.5 (gpt-3.5-turbo).
    \item {\bf GPT-4}~\citep{OpenAI2023GPT4TR}: is a large multimodal model, achieving human-level performance on various NLP benchmarks. 
    \item {\bf Self-negotiation}: The same LLM acts as both the roles of the generator and the discriminator.
\end{tightitemize}

\subsection{Results and analysis}

Experiment results are shown in Table \ref{tab:main-results}. As can be seen in the table, compared to vanilla ICL, following the generate-discriminate paradigm with one LLM (self-negotiation) receives performance gains on six sentiment analysis datasets: GPT-3.5 gains +0.9 on average;  GPT-4 receives +1.0 acc on average. This phenomenon illustrates that the LLM, performing as the answer discriminator, can correct a portion of errors caused by  the task generator.  

We also observe that using two different LLMs as the task generator and task discriminator in turn introduces significant performance improvements compared to merely using one model. Negotiations with two LLMs outperform the self-negotiation method by +1.7, +2.1, and +2.3 in terms of accuracy on MR, Twitter, IMDB datasets, respectively. The reason for this phenomenon is that using two different LLMs finish the sentiment analysis task through negotiations can take the advantage of different understandings of the given input and unleash the power of two LLMs, leading to more accurate decisions. 

We also find that when introduce a third LLM to resolve the disagreement between the flippled-roled negotiations, 
additional performance boost can be obtained.  
This demonstrates that the third LLM can resolve conflicts between two LLMs through multiple negotiations and improve performances on the sentiment analysis task. It is noteworthy that the  multi-model negotiation method outperforms the supervised method RoBERTa-Large by +0.9 on the MR dataset, and bridges the gap between vanilla ICL and the supervised method: achieving 
94.1 (+1.4) accuracy on SST-2; 92.1 (+2.7) on Twitter; 96.3 (+2.5) on Yelp-2; 87.2 (+3.7) on Amazon-2; and 94.5 (+6.0) on IMDB dataset.
\section{Ablation Studies}
In this section, we perform ablation studies on the Twitter dataset to better understand the mechanism behind the negotiation framework.

\subsection{Who takes which role matters}
In the negotiation framework, there are two roles, the generator and the discriminator, which two separate LLMs take. 
Table \ref{flip} shows the performance for setups where  GPT-3.5 and GPT-4 take different roles. 

As can be seen, when GPT-3.5 acts as the generator, and GPT-4 acts as the discriminator (G3.5-D4 for short), the performance (68.8) is better than 
single GPT-3.5 without negotiation (65.2), but worse than 
single GPT-4 without negotiation (69.5). 
In contrast, negotiation-based configurations with GPT-4 acting as the generator (G4-D3.5 and G4-D4) consistently outperforms standalone GPT-4 or GPT-3.5 models without negotiation. 
These results underscore the pivotal role that the generator plays in influencing the negotiation outcome. Furthermore, we observe G4-D3.5 can beat G4-D4. We attribute such advantage to the hypothesis that utilizing heterogeneous LLMs for distinct roles could optimize the negotiation's performance.

\begin{table}
\center
\begin{tabular}{lll}\toprule
G & D &ACC \\\hline
GPT-3.5 &- & 65.2 \\
 GPT-4 & -&69.5 \\
GPT-3.5 &GPT-3.5  &66.8\\
GPT-3.5 &GPT-4  &65.2\\
GPT-4 &GPT-3.5  &72.8\\
GPT-4 & GPT-4  &72.2\\
\hline
\end{tabular}
\caption{Performance on the Twitter dataset with GPT-3.5 and GPT-4 taking different roles. G denotes generator and D denotes discriminator.}
\label{flip}
\end{table}

\subsection{Consensus Percentage}
\begin{table}
\center
\begin{tabular}{lcc}\hline
&G3.5-D4 & G4-D3.5\\\hline
2 turns agree&65\%&76\%\\
3 turns agree&29\%&21\%\\
3 turns disagree&6\%&3\%\\\hline
\end{tabular}
\caption{Consensus percentage for different setups on the Twitter dataset. G3.5-D4 denotes GPT-3.5 acts as the generator and GPT-4 acts as the discriminator.}
\label{consensus}
\end{table}
Table \ref{consensus}
consensus percentage for different setups. 
As can be seen, when GPT-4 acts as the generator, the negotiation is more likely to reach a consensus, 
or reach a consensus in fewer turns. 
The explanation is intuitive:
for the twitter task, we can see from table \ref{tab:main-results} that GPT-4 obtains better performances that GPT-3.5, which means the reasoning process
for GPT-4 is more sensible than 3.5, making the decision of the former more likely to be agreed on. 

\subsection{Effect of the Reasoning Process}
In the negotiation process, LLMs are asked to articulate the reason process, a strategy akin to CoT\citep{wei2022chain}. 
We examine the importance for listing reasons in negotiation by removing the reasoning process and asking LLMs to only output  
decisions.
Results are shown in Table \ref{reason}.
As can be seen, for the three setups, single GPT-3.5, where only GPT-3.5 is used without negotiation, 
single GPT-4, where only GPT-4 is used without negotiation, and GPT-3.5+GPT-4 where negotiation is employed, 
performances all degrade when the reasoning process is removed.
But interestingly, we see a greater degrade (-2.3) for the negotiation than the single model setup (-1.2 for single-GPT-3.5 and -0.9 for single-GPT-4).
This is in accord with our expectation as 
 the reasoning process is of greater significance in the negotiation setup.

\begin{table}
\center
\begin{tabular}{lll}\hline
Model &Reason&ACC \\\hline
single GPT-3.5  &w& 65.2 \\
single GPT-3.5  &wo& 64.0 (-1.2) \\\hline
single GPT-4  &w& 69.5 \\
single GPT-4  &wo& 68.6 (-0.9) \\\hline
GPT-3.5+GPT-4 &w & 74.6 \\
GPT-3.5+GPT-4 &wo & 72.3 (-2.3) \\\hline
\end{tabular}
\caption{Effect of removing the reasoning process on the Twitter dataset.}
\label{reason}
\end{table}

\section{Conclusion}
In this paper, we investigate the limitations of singular LLM-based sentiment analysis methods and introduce a novel role-flipping multi-LLM negotiation method to enhance both the accuracy and interpretability of sentiment categorizations. Empirical findings on multiple benchmarks show
the superiority of our approach compared to
traditional ICL and many supervised methods. Future work could explore optimizing the framework for speed and resource consumption, adapting the underlying principles to other NLP tasks, and designing explicit negotiation modules that identify and mitigate the impact of biases and decoding errors present in individual LLMs.

\bibliography{custom}
\bibliographystyle{acl_natbib}

\end{document}

%% file: header.tex
\usepackage{tikz-dependency}
\usepackage{booktabs}
\usepackage{comment}
\usepackage{graphicx} 
\usepackage{amssymb}
\usepackage{times}
\usepackage{latexsym}
\usepackage{pifont}
\usepackage{times}
\usepackage{latexsym}
\usepackage{comment}

\usepackage{url}
\usepackage{amsmath}

\usepackage{CJKutf8}
\usepackage[boxed]{algorithm}
\usepackage{caption}
\usepackage{algpseudocode}

\usepackage[T1]{fontenc}
\usepackage[utf8]{inputenc}

\usepackage{microtype}
\usepackage{tcolorbox}
\usepackage{tikz,pgfplots}
\usepgfplotslibrary{groupplots}
\usepackage{nicefrac} 
\pgfplotsset{compat=1.13}

\usepackage{enumerate}
\usepackage{amssymb}

\newcommand{\todo}[1]{\iffalse #1 \fi {\textcolor{blue}{ \textbf{[TODO]} }}}

\definecolor{g-red}{HTML}{DB4437}
\definecolor{g-blue}{HTML}{4285F4}
\definecolor{g-green}{HTML}{0F9D58}
\definecolor{g-yellow}{HTML}{F4B400}
\definecolor{g-orange}{HTML}{FF9800}
\definecolor{g-grey}{HTML}{9E9E9E}
\definecolor{shannon}{HTML}{304FFE}
\definecolor{uw}{RGB}{138,43,226}
\definecolor{stanford}{RGB}{255,69,0}
\definecolor{const}{RGB}{68, 110, 182}
\definecolor{head}{RGB}{246, 180, 32}
\definecolor{freq}{RGB}{0, 0, 0}
\definecolor{ao}{rgb}{0.0, 0.5, 0.0}
\definecolor{asparagus}{rgb}{0.53, 0.66, 0.42}
\definecolor{amber}{rgb}{1.0, 0.49, 0.0}
\definecolor{alizarin}{rgb}{0.82, 0.1, 0.26}
\definecolor{applegreen}{rgb}{0.55, 0.71, 0.0}
\definecolor{amethyst}{rgb}{0.6, 0.4, 0.8}
\definecolor{auburn}{rgb}{0.43, 0.21, 0.1}
\definecolor{positive}{HTML}{dc3b96}
\definecolor{rationale}{HTML}{2d8cfa}
\definecolor{demo}{HTML}{c375d5}
\definecolor{discriminator}{HTML}{ff8e57}
\definecolor{response}{HTML}{9a9b72}
\definecolor{negative}{HTML}{27a723}

\newenvironment{tightitemize}%
  {\begin{itemize} %
    \setlength{\itemsep}{1.2pt}%
    \setlength{\parskip}{0.5pt}%
    }%
  {\end{itemize}}

\usepackage{caption}
\usepackage{booktabs}

\usepackage{babel}

\newcommand{\heart}{\text{\small \ding{170}}}
\usepackage{multirow}
\usepackage{subcaption}
\usepackage{booktabs} 
\usepackage{tablefootnote} 
\usepackage{threeparttable}
\usepackage{graphicx}
\usepackage{adjustbox}
\usepackage[referable]{threeparttablex}
\usepackage{multicol}
\usepackage{booktabs}

%% file: math_commands.tex
\usepackage{amsmath,amsfonts,bm}

\def\eqref#1{equation~\ref{#1}}

\def\1{\bm{1}}

\DeclareMathAlphabet{\mathsfit}{\encodingdefault}{\sfdefault}{m}{sl}
\SetMathAlphabet{\mathsfit}{bold}{\encodingdefault}{\sfdefault}{bx}{n}



%% file: tables/main_result.tex
\begin{table*}[t]
\small
     \scalebox{0.9}{
      \begin{tabular}{lccccccc}
      \toprule
       &SST-2 & Movie Review & Twitter &	Yelp-Binary	& Amazon-Binary	& IMDB & {\bf Average}
       \\\midrule
      \multicolumn{8}{c}{\underline{\tt \bf Supervised Methods}}\\\\
      DRNN~\citep{wang2018disconnected} & - & 90.4 & - & 97.3  & 96.4 & 95.3 & - \\
      RoBERTa~\citep{liu2019roberta} & 96.0&	91.2&	71.4&	98.6	&96.0	&95.9 & 91.5  \\
      XLNet~\citep{yang2019xlnet} & {\bf 97.0\S} & - & - & {\bf 98.6\S}  & {\bf 97.9\S} & {\bf 96.2\S} & - \\
      UDA~\citep{xie2020unsupervised}& - & - & - & 97.9  & 96.5 & 95.8 & - \\
      BERTweet~\citep{nguyen2020bertweet} & - & - & {\bf 71.6\S} & -  & - & - & - \\
      EFL~\citep{wang2021entailment} & 96.9 & {\bf 92.5\S} & - & -  & - & 96.1 & - \\
      \midrule
      \multicolumn{8}{c}{\underline{\tt \bf LLM ICL Baselines}}\\\\
      InstructGPT3.5~\citep{ouyang2022training}&92.4 & 89.6 & - & - & - & 90.7 & - \\
      \citet{zhang2023sentiment} &  &  & &  &  & &  \\
      \phantom{a} - w/ Flan-UL2& 97.4 & 93.8 & 47.9 & - & - & - & - \\
      \phantom{a} - w/ T5& 91.4 & 85.7 & 53.2 & 92.4 & - & 90.0 & - \\
      \phantom{a} - w/ GPT-3.5& 95.3 & 90.2 & 64.3 & - & - & - & - \\
      IDS~\citep{Qin2023InContextLW} & 95.8 & - & - & 94.2 & 95.7 & - & - \\
      GPT-4~\citep{OpenAI2023GPT4TR} & 92.5 & - & - & 94.2 & - & - & - \\
      \midrule
      \multicolumn{8}{c}{\underline{\it \bf Our Implementation }}\\
      \multicolumn{8}{l}{\it \bf Vanilla ICL}\vspace{1pt}\\
      \phantom{a} - w/ GPT-3.5 &92.7&	90.2&	65.2&	93.8&	84.8	& 90.6 & 86.2   \\
      \phantom{a} - w/ GPT-4 &93.2	&89.4&	69.5&	95.2&	83.5&	88.5 & 86.6  \\\midrule
      \multicolumn{8}{l}{\it \bf Self-Negotiation}\vspace{1pt}\\
      \phantom{a} - w/ GPT-3.5 &93.2&	90.6	&66.8	&94.5	&86.0	&91.7 & 87.1  \\
      \phantom{a} - w/ GPT-4 &93.3	&90.3&	72.2&	95.5&	84.3	& 89.7& 87.6   \\
    \multicolumn{8}{l}{\it \bf Negotiation with LLMs}\vspace{1pt}\\
    \phantom{a} - w/ GPT-3.5+GPT-4&93.8&	92.3&	74.3&	96.3	&86.9	&94.0&89.6   \\
    \phantom{a} - w/ GPT-3.5+GPT-4+InstructGPT3.5&94.1&	92.7	&74.6	&96.3	&87.2	&94.5&89.8  \\
      \bottomrule
    \end{tabular}
    }
    \caption{Accuracy performances of different settings on benchmarks. Performances with \S~ denote current state-of-the-art.
    }
    \label{tab:main-results}
    \end{table*}